\documentclass[10pt,twocolumn,letterpaper]{article}
\usepackage[accsupp]{axessibility}
\usepackage{iccv}              
\usepackage{booktabs}
\usepackage{multirow}
\usepackage{makecell}
\definecolor{iccvblue}{rgb}{0.21,0.49,0.74}
\usepackage[pagebackref,breaklinks,colorlinks,allcolors=iccvblue]{hyperref}

\def\paperID{3731} 
\def\confName{ICCV}
\def\confYear{2025}

\newcommand{\subtitle}[1]{{\noindent}{\textbf{#1}}}

\title{Learning to See in the Extremely Dark}

\author{
    Hai Jiang\textsuperscript{\rm1,\thanks{Equal contribution}}, Binhao Guan\textsuperscript{\rm2,\footnotemark[1]}, Zhen Liu\textsuperscript{\rm2}, Xiaohong Liu\textsuperscript{\rm3}, Jian Yu\textsuperscript{\rm4}, Zheng Liu\textsuperscript{\rm4}, \\ 
    Songchen Han\textsuperscript{\rm1}, Shuaicheng Liu\textsuperscript{\rm2,\thanks{Corresponding author}} \\
    \textsuperscript{\rm1}School of Aeronautics and Astronautics, Sichuan University \\ 
    \textsuperscript{\rm2}University of Electronic Science and Technology of China \\
    \textsuperscript{\rm3}Shanghai Jiao Tong University \textsuperscript{\rm4}National Innovation Center for UHD Video Technology \\
    {\tt\small jianghai@stu.scu.edu.cn, \{guanbinhao@std., liushuaicheng@\}uestc.edu.cn}
}

\begin{document}
\maketitle
\begin{abstract}
    Learning-based methods have made promising advances in low-light RAW image enhancement, while their capability to extremely dark scenes where the environmental illuminance drops as low as 0.0001 lux remains to be explored due to the lack of corresponding datasets. To this end, we propose a paired-to-paired data synthesis pipeline capable of generating well-calibrated extremely low-light RAW images at three precise illuminance ranges of 0.01-0.1 lux, 0.001-0.01 lux, and 0.0001-0.001 lux, together with high-quality sRGB references to comprise a large-scale paired dataset named \textbf{S}ee-\textbf{i}n-the-\textbf{E}xtremely-\textbf{D}ark (\textbf{SIED}) to benchmark low-light RAW image enhancement approaches. Furthermore, we propose a diffusion-based framework that leverages the generative ability and intrinsic denoising property of diffusion models to restore visually pleasing results from extremely low-SNR RAW inputs, in which an Adaptive Illumination Correction Module (AICM) and a color consistency loss are introduced to ensure accurate exposure correction and color restoration. Extensive experiments on the proposed SIED and publicly available benchmarks demonstrate the effectiveness of our method. The code and dataset are available at \url{https://github.com/JianghaiSCU/SIED}.
\end{abstract}

\section{Introduction}\label{sec: intro}
Restoring high-quality sharp images from low-light observations is a challenging task, as it requires improving global and local contrast, suppressing amplified noise, and preserving details, which is critical for realistic applications such as night-time photography and surveillance. Recent advances in deep learning-based low-light image enhancement (LLIE) have significantly propelled this field forward. Generally, learning-based methods can be divided into two categories in terms of their input format: sRGB-based methods~\cite{RetinexNet, LOLV2, UHD_ICLR, KinD++, GCCIM, Bread, Diff-Retinex, GSAD, PyDiff, Retinexformer, DiffLL} and RAW-based methods~\cite{SID, DID, LLPackNet, RRT, DNF, MCR, RAWMamba, Exposurediffusion, LDM-ISP}. Compared to the extensively studied sRGB-based LLIE methods, RAW-based approaches have garnered growing attention due to their ability to exploit more informative low-intensity RAW signals and superior noise modeling capabilities.
\begin{figure}[!t]
    \centering
    \includegraphics[width=\linewidth]{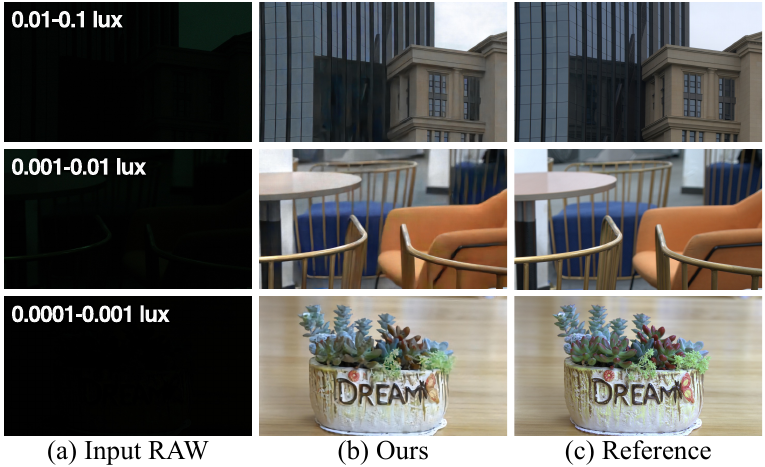}
    \caption{An illustration of our synthesized low-light RAW images at three extremely low illuminance levels of 0.01-0.1 lux, 0.001-0.01 lux, and 0.0001-0.001 lux are shown in (a), with the restored results from our method in (b) and the reference high-quality sRGB images in (c). The input low-light RAW images are visualized by a simple demosaicing process.}
    \label{fig: teaser}
\end{figure}

To this date, numerous efforts have concentrated on transforming low-light noisy RAW images into normal-light clean sRGB images, employing either single-stage or multi-stage training strategies. Single-stage methods~\cite{SID, DID, SGN, LLPackNet, RRT} attempt to model both the noisy-to-clean and RAW-to-sRGB transformations within a single network. However, the hybrid mappings across these two distinct domains inevitably mislead the enhancement process, causing domain ambiguity issues~\cite{DNF}. Furthermore, the complexity of multiple nonlinear processes in the RAW-to-sRGB transformation exacerbates the problem, often resulting in color distortion and blurry details. To overcome these limitations, multi-stage methods~\cite{EEMEFN, RRENet, LDC, MCR, DNF, RAWMamba} decouple the tasks using intermediate supervision across distinct domains and fully exploit the unique characteristics of RAW and sRGB formats to achieve improved results.

On the other hand, qualified datasets are indispensable for advancing learning-based low-light RAW enhancement methods. Existing datasets~\cite{SID, MCR, SDSD, SMID} typically collect paired images in dark environments via varying camera parameters, where the normal-light reference images are obtained by extending exposure times. However, due to the influence of environmental light sources and brightness fluctuations during data collection, these datasets offer only coarse illuminance levels for the captured low-light RAW images (e.g., 0.2-5.0 lux for indoor scenes and 0.03-0.3 lux for outdoor scenes in the SID~\cite{SID} dataset). Moreover, in truly extreme dark conditions, such as environments with illuminance as low as 0.0001 lux, obtaining well-exposed reference images through long exposure time is unattainable, which would result in residual noise and motion blur.

To this end, we propose a paired-to-paired data synthesis pipeline to prepare a more challenging dataset specifically designed for extremely low-light RAW image enhancement, named \textbf{S}ee-\textbf{i}n-the-\textbf{E}xtremely-\textbf{D}ark (\textbf{SIED}), which generates low-light RAW images at three well-calibrated extremely low illuminance levels, along with sharp sRGB references, through three aspects: 1) collecting qualified low-light RAW images with three precise illuminance ranges of 0.01-0.1 lux, 0.001-0.01 lux, and 0.0001-0.001 lux in a professional optical laboratory, 2) capturing paired low-light and normal-light images across diverse real-world scenes by adjusting the camera parameters in line with previous data collection strategies~\cite{SID, MCR} where the normal-light images serve as high-quality references, 3) mapping the illumination of the captured low-light images to the above three distinct illuminance ranges to align with the standard data collected in the laboratory and employing the calibrated sensor noise at each range to the adjusted low-light images to emulate realistic dark environments. With professional laboratory calibration and our paired-to-paired data synthesis strategy, our pipeline ensures realistic low-light RAW inputs for extremely dark scenes and corresponding high-quality sRGB references, as shown in Fig.~\ref{fig: teaser}(a) and (c).

In addition to the dataset, we propose a diffusion-based framework trained with a multi-stage strategy, which leverages the generative ability and intrinsic denoising capability of diffusion models for extremely low-light RAW image enhancement. Specifically, we first convert the paired low-light RAW image and normal-light sRGB image into latent space, where the encoded RAW feature is sent to the designed Adaptive Illumination Correction Module (AICM) for exposure correction, which differs from the previous methods~\cite{SID, DID, SGN, DNF, RAWMamba} employing the exposure information of reference images to perform pre-amplification, aiming to achieve better restoration results and avoid the exposure bias in the following diffusion processes. Subsequently, the encoded sRGB feature serves as input for the diffusion model for restoration with the guidance of the amplified RAW feature, in which a color consistency loss is further proposed to facilitate the diffusion model to generate reconstructed sRGB features with accurate color mapping. As shown in Fig.~\ref{fig: teaser}(b), our method effectively improves global and local contrast, presents vivid color, and avoids noise amplification, resulting in visually satisfactory images in extremely dark environments. 

To summarize, our main contributions are as follows:
\begin{itemize}
  \item We introduce a novel paired-to-paired data synthesis pipeline capable of generating low-light RAW images at three distinct illuminance levels, along with high-quality sRGB images, to form the large-scale SIED dataset for extremely low-light RAW image enhancement.
  \item We propose a diffusion-based framework that leverages the generative capability and intrinsic denoising property of diffusion models to restore visually pleasing sharp images from extremely low-light RAW inputs.
  \item Extensive experiments show that our method outperforms existing state-of-the-art single-stage and multi-stage competitors both qualitatively and quantitatively.
\end{itemize}

\section{Related Work}\label{sec: related_work}
\subtitle{Low-light RAW Image Enhancement methods.} RAW images in comparison to sRGB images are more informative and have thus been widely used for image enhancement in low-light conditions. With the development of deep learning, numerous efforts have been made to transform the low-light RAW domain into the sharp sRGB domain with deep neural networks, which can be divided into single-stage and multi-stage methods. Single-stage methods~\cite{SID, DID, SGN, LLPackNet, RRT} typically promote the network to generate high-quality sRGB images from input RAW images through well-designed optimization objectives directly. However, it is difficult to learn multiple nonlinear transformations with a single network, making them often insufficient for accurate detail reconstruction and color mapping. Multi-stage approaches~\cite{EEMEFN, LDC, MCR, RRENet, DNF, RAWMamba} are designed to overcome the above limitations, achieving improved results by decoupling tasks and thereby effectively reducing ambiguities between different domains. 
\begin{figure*}[!t]
    \centering
    \includegraphics[width=\linewidth]{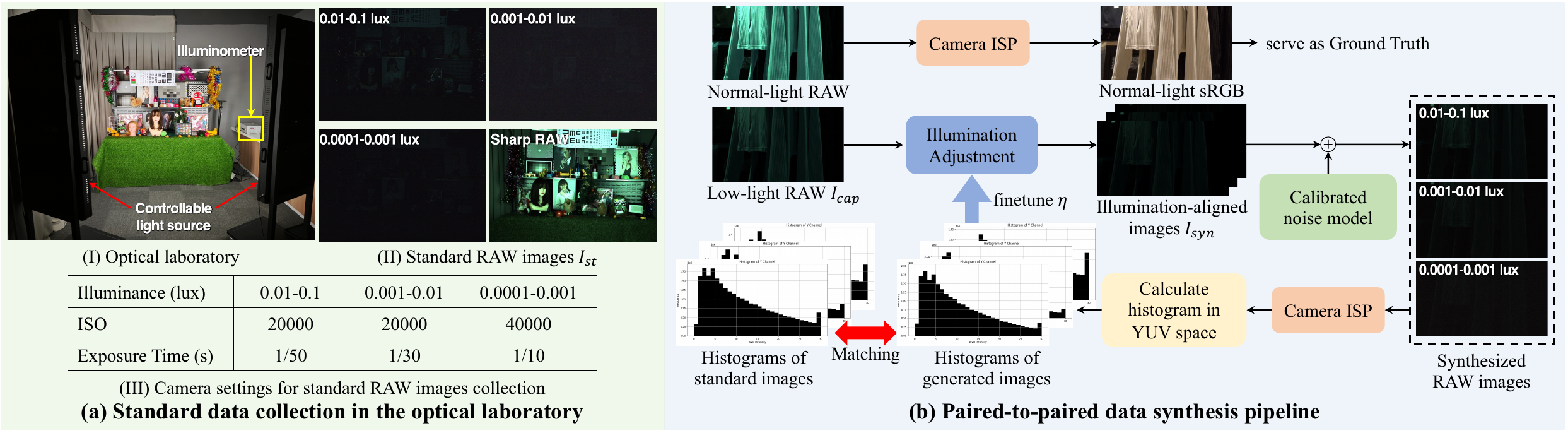}
    \caption{(a) illustrates the optical laboratory, collected standard RAW images, and camera settings adopted for image capturing. (b) presents our proposed paired-to-paired data synthesis strategy. The RAW images are visualized by a simple demosaicing process.}
    \label{fig: synthesis_pipeline}
\end{figure*}

\subtitle{Low-light RAW Image Datasets.} High-quality paired datasets are indispensable for advancing learning-based methods. The SID~\cite{SID} dataset contains paired low-/normal-light images captured by Sony and Fuji cameras, where the environmental illuminance is 0.2-5.0 lux for indoor scenes and 0.03-0.3 lux for outdoors. The SDSD dataset~\cite{SDSD} provides spatially aligned video pairs from dynamic scenes for low-light RAW video enhancement, with the illuminance of approximately 0.8–2.0 lux. The SMID dataset~\cite{SMID} includes static videos with ground truth and dynamic videos without ground truth, where most scenes are in the 0.5-5.0 lux range. Due to the environmental complexity and variations in data collection strategies, the above datasets only provide rough illuminance ranges and lack data captured in darker scenes. In this paper, we design a new paired-to-paired data synthesis pipeline to prepare a more challenging dataset that contains low-light RAW images with three well-calibrated illuminance ranges as low as 0.0001 lux.

\section{See-in-the-Extremely-Dark Dataset}\label{sec: sied_dataset}
To generate low-light RAW images with precise illumination ranges, we first use Sony \(\alpha\)7R\text{III} and Canon EOS R to capture qualified low-light RAW images with illumination ranges of 0.01-0.1 lux, 0.001-0.01 lux, and 0.0001-0.001 lux in an optical laboratory with controllable light sources and a professional illuminometer (PHOTO-2000\(\mu\)), as shown in Fig.~\ref{fig: synthesis_pipeline}(a). Subsequently, instead of synthesizing low-light images from normal-light images as in the previous methods, we propose a paired-to-paired synthesis strategy to enable the synthesized images to approximate realistic images, as shown in Fig.~\ref{fig: synthesis_pipeline}(b), which consists of three steps: 1) collecting paired low-light RAW and sharp sRGB images with the above two cameras in various scenes, 2) manually adjusting the illumination of the low-light images to produce images with specific illuminance levels that align with the standard laboratory data, 3) adding the calibrated noise model to satisfy the realistic dark environments.

\subtitle{Realistic Paired Data Collection.} We collect paired low-light RAW images and normal-light sRGB images across various real-world static scenes by varying the camera parameters following~\cite{SID, SDSD}, where the exposure times for reference images range between 1/10 and 1/200 second, which is 20 to 200 times longer than for low-light images. In each scene, we adapt camera settings such as aperture, ISO, and focal length to maximize the quality of the reference image and fix the settings when capturing the low-light image. To ensure content consistency, we adopt tripods to mount the cameras with remote applications to control exposure settings and capture. To improve the scene richness of the dataset, we crop the original full-resolution images, i.e., 7,952$\times$5,304 for Sony and 6,720$\times$4,480 for Canon, to images with 3,840$\times$2,160 resolution, resulting in 1,680 paired images in each subset for subsequent synthesis.
\begin{figure*}[!t]
    \centering
    \includegraphics[width=\linewidth]{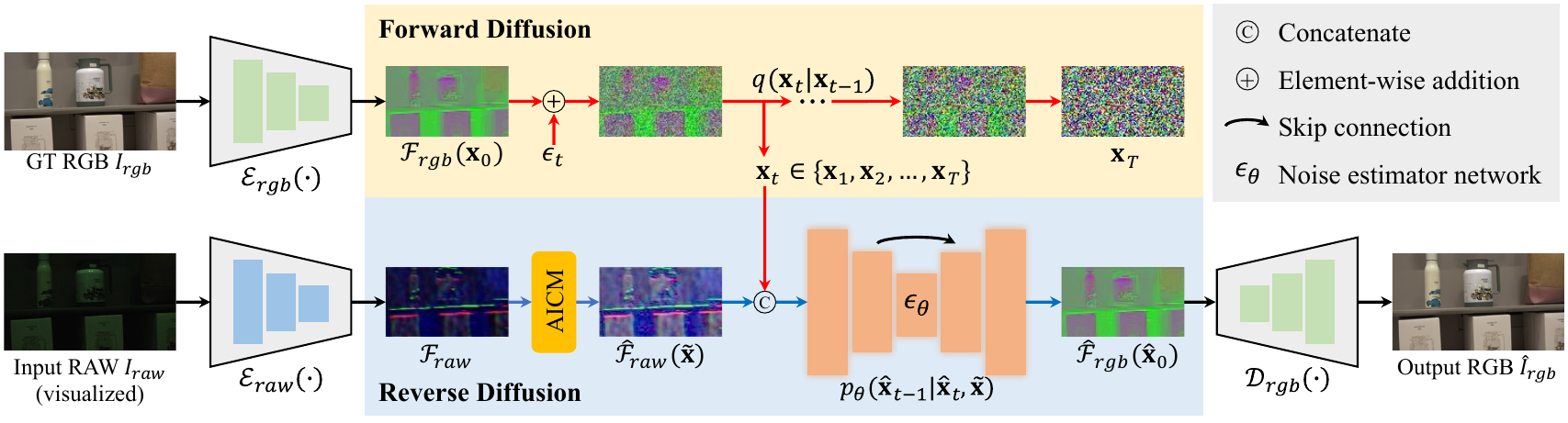}
    \caption{The overall pipeline of our proposed framework. We first employ a RAW encoder $\mathcal{E}_{raw}(\cdot)$ and a sRGB encoder $\mathcal{E}_{rgb}(\cdot)$ to convert the input noisy RAW image $I_{raw}$ and reference GT image $I_{rgb}$ into latent space denoted as $\mathcal{F}_{raw}$ and $\mathcal{F}_{rgb}$. Then, the encoded sRGB feature is taken as the input of the diffusion model to perform the forward diffusion process, while the encoded RAW feature is sent to the proposed Adaptive Illumination Correction Module (AICM) to generate contrast improved feature denoted as $\hat{\mathcal{F}}_{raw}$, aiming to avoid the exposure bias in diffusion processes. In the reverse diffusion process, the refined RAW feature $\hat{\mathcal{F}}_{raw}$ serves as guidance to generate the reconstructed sRGB feature $\hat{\mathcal{F}}_{rgb}$ from the noised tensor $\mathbf{x}_{t}$, which is replaced by randomly sampled Gaussian noise $\hat{\mathbf{x}}_{T}$ during inference. Finally, the reconstructed sRGB feature is sent to the sRGB decoder $\mathcal{D}_{rgb}(\cdot)$ to produce the final restored result $\hat{I}_{rgb}$.    
    }
    \label{fig: Pipeline}
\end{figure*}

\begin{table}[!t]
  \centering
    \caption{The KL divergence ($\downarrow$) of illumination histogram distributions between our synthesized images and laboratory images.}
   \resizebox{0.9\linewidth}{!}{
    \begin{tabular}{l|ccc}
    \toprule
    & 0.01-0.1 lux & 0.001-0.01 lux & 0.0001-0.001 lux \\
    \midrule
    Canon & 0.017 &	0.018 & 0.011 \\
    Sony  & 0.011 &	0.009 &	0.059 \\
    \bottomrule
    \end{tabular}}
  \label{tab: kl_dataset}%
\end{table}%

\subtitle{Illumination Alignment.} Since the illumination information in RAW images is linearly correlated with the photon intensity, we employ the exposure of standard data $I_{st}$ to adapt the captured low-light images $I_{cap}$, which is similar to the amplification strategy in previous methods~\cite{SID, DNF, RAWMamba}, to simulate the realistic illumination degradation as:
\begin{equation}\label{eq: exposure_adjust}
    I_{syn} = I_{cap} * (\frac{\operatorname{Expo}(I_{st})}{\operatorname{Expo}(I_{cap})} + \eta),
\end{equation}
where $I_{syn}$ is the synthesized image aligned with the standard data across three illuminance ranges, $\operatorname{Expo}(\cdot)$ calculates the mean value of Bayer channels to characterize illuminance information, and $\eta$ is the manually defined factor.

\subtitle{Noise Addition.} In realistic dark scenes, noise is the inevitable degradation factor. To generate more realistic low-light RAW images, we first estimate the sensor noise model of Canon and Sony cameras in the optical laboratory, fitting the Gaussian and Poisson noise distributions under various ISO values. Moreover, as mentioned in~\cite{ELD, dark_frame}, the noise model under extremely low-light conditions should not be considered a pure P+G model. Therefore, we add the dark-frame database used for calibrating Gaussian noise to complement the noise types that are hard to model explicitly to better satisfy the characteristics of realistic noise distribution~\cite{dark_frame}, resulting in the calibrated noise model is the combination of the Gaussian, Poisson, and dark-frame distribution. Finally, we adopt the ISO-dependent noise addition strategy~\cite{noise_modeling} to apply the calibrated noise model within an equivalent ISO range between 100 and 20,000 to the illumination-aligned images with the illuminance of 0.01-0.1 lux and 0.001-0.01 lux, while the ISO range is set between 100 and 40,000 for 0.0001-0.001 lux.
\begin{figure}[!t]
    \centering
    \includegraphics[width=\linewidth]{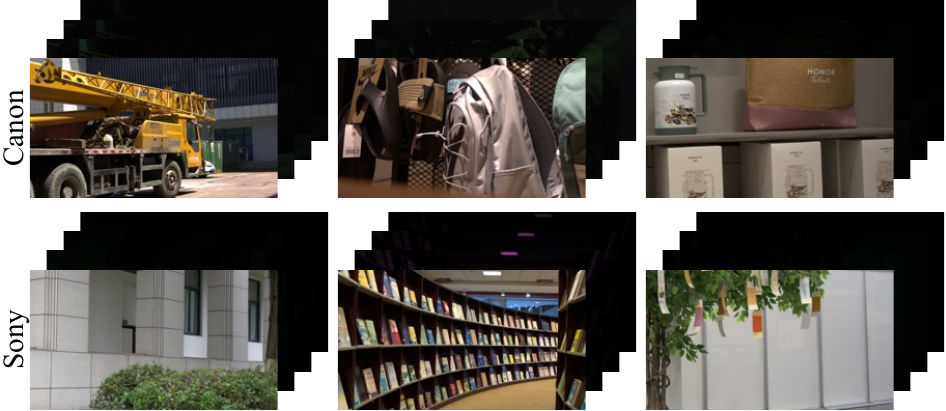}
    \caption{Examples from the Canon and Sony subsets of our SIED dataset, with reference sRGB images shown in front and low-light RAW images at three illuminance levels shown in the back.}
    \label{fig: dataset}
\end{figure}

To refine the illumination of generated images, we adopt a fixed ISP pipeline to transform the generated RAW image and standard images into YUV space, and manually finetune $\eta$ to match the illumination histograms of the two images in the Y channel, which represents the illumination information of images~\cite{Bread}. As shown in Table~\ref{tab: kl_dataset}, the mean KL divergence of the histogram distributions between our generated images and standard laboratory images is less than 0.06 across three illuminance levels. Overall, the synthesized low-light noisy RAW images with three relatively precise illuminance ranges and the corresponding normal-light sRGB images comprise each sample of our See-in-the-Extremely-Dark (SIED) dataset, containing 1,680 paired images in each illuminance level of Sony and Canon subsets where 1,500 pairs are selected for training and 180 pairs are used for evaluation. We present several samples in Fig.~\ref{fig: dataset} and more details can be found in the supplementary.

\section{Methodology}\label{sec: method}
\subsection{Overview}\label{subsec: overview}
Low-light RAW image enhancement presents several critical concerns: contrast enhancement, noise suppression, detail reconstruction, and global color mapping. To overcome these challenges, we propose a diffusion-based framework that leverages the generative ability of diffusion models to achieve visually satisfactory results, as illustrated in Fig.~\ref{fig: Pipeline}. Given a paired low-light RAW image $I_{raw} \in \mathbb{R}^{H\times W \times 1}$ and reference sRGB image $I_{rgb}  \in \mathbb{R}^{H\times W \times 3}$, we first employ a RAW encoder $\mathcal{E}_{raw}(\cdot)$ and a sRGB encoder $\mathcal{E}_{rgb}(\cdot)$, to transform the input images into latent space denoted as $\mathcal{F}_{raw} \in \mathbb{R}^{h\times w \times C}$ and $\mathcal{F}_{rgb} \in \mathbb{R}^{h\times w \times C}$. Then, we design an Adaptive Illumination Correction Module (AICM) to improve the contrast of the encoded RAW feature denoted as $\hat{\mathcal{F}}_{raw}$, aiming to avoid the exposure bias in the following diffusion processes and achieve better restoration results. Subsequently, the encoded sRGB feature serves as the input for the diffusion model with the guidance of the amplified RAW feature to generate restored feature $\hat{\mathcal{F}}_{rgb}$. Finally, the restored feature will be sent to a sRGB decoder $\mathcal{D}_{rgb}(\cdot)$ for reconstruction to produce the final restored image $\hat{I}_{rgb}$.
\begin{figure}[!t]
    \centering
    \includegraphics[width=\linewidth]{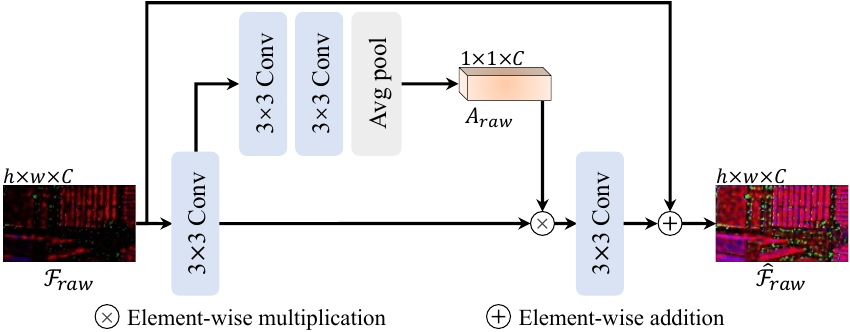}
    \caption{The detailed architecture of our proposed AICM.}
    \label{fig: AICM}
\end{figure}

\subsection{Adaptive Illumination Correction Module}\label{subsec: AICM}
Previous methods~\cite{SID, DID, EEMEFN, DNF, RAWMamba} have shown that pre-amplifying low-light images can lead to improved restoration results. However, most approaches rely on the exposure values of reference normal-light images as a prior for amplification, which is unlikely to be available in practical applications. Moreover, while diffusion models demonstrate impressive generative capabilities, low-frequency generative biases would be encountered, particularly in exposure~\cite{diffsuion_bias}. 

To this end, we propose a simple yet effective Adaptive Illumination Correction Module (AICM), which estimates the amplification factor from the low-light RAW feature to perform illumination correction in the latent space, aiming to avoid exposure bias in diffusion processes and obtain better restoration results. As shown in Fig.~\ref{fig: AICM}, we first use a convolutional layer to embed the input feature $\mathcal{F}_{raw}$ followed by cascaded convolutional layers with an adaptive average pooling layer to estimate the per-channel amplification coefficients $A_{raw} \in \mathbb{R}^{1\times 1 \times C}$, which is used to improve the contrast of the embedded feature. Finally, the refined feature is reconstructed by a convolutional layer with residual content to generate the amplified feature $\hat{\mathcal{F}}_{raw}$. 

To enable adaptive exposure correction in AICM and ensure the amplified RAW feature shares content consistency with the input feature, we propose an illumination correction loss $\mathcal{L}_{icl}$ based on the Retinex theory~\cite{Retinex} as:
\begin{equation}
    \mathcal{L}_{icl} = \|\mathbf{L}_{\hat{\mathcal{F}}_{raw}} - \mathbf{L}_{\tilde{\mathcal{F}}_{raw}}\|_{1} + \|\mathbf{R}_{\hat{\mathcal{F}}_{raw}} - \mathbf{R}_{\mathcal{F}_{raw}}\|_{1},
\end{equation}
where the $\tilde{\mathcal{F}}_{raw}$ is the encoded feature of the normal-light RAW image. $\mathbf{L}$ and $\mathbf{R}$ represent the illumination map and the reflectance map, respectively, which are obtained through the decomposition strategy in~\cite{LIME}.

\subsection{Diffusion-based RAW-to-sRGB Reconstruction}\label{subsec: diffusion_model}
The main concern after obtaining the amplified feature is to convert it into a high-quality sRGB feature. To achieve this, we propose to leverage the generative ability and intrinsic denoising capability of diffusion models to facilitate the reconstruction from noisy RAW data to clean sRGB representation. Our approach follows standard diffusion models~\cite{ddpm, conditional_ddpm, ddim} that perform forward diffusion and reverse diffusion processes to generate reconstructed results. 

\subtitle{Forward Diffusion.} The encoded sRGB feature $\mathcal{F}_{rgb}$ is taken as the input of the diffusion model, denoted as $\mathbf{x}_{0}$, to perform the forward diffusion process, in which a predefined variance schedule $\{\beta_{1},\beta_{2},...,\beta_{T}\}$ is employed to progressively convert $\mathbf{x}_{0}$ into Gaussian noise $\mathbf{x}_{T}\sim{\mathcal{N}(\mathbf{0, I})}$ over $T$ steps, which can be formulated as: 
\begin{equation}\label{eq: forward_diffusion}
    q(\mathbf{x}_{t}|\mathbf{x}_{t-1})=\mathcal{N}(\mathbf{x}_{t};\sqrt{1-\beta_{t}\mathbf{x}_{t-1}}, \beta_{t}\mathbf{I}),
\end{equation}
where $\mathbf{x}_{t}$ indicates the corrupted noisy data at time-step $t\in[0,T]$. With parameter renormalization, we can obtain $\mathbf{x}_{t}$ directly from the input $\mathbf{x}_{0}$ and thereby simplify Eq.(\ref{eq: forward_diffusion}) into a closed expression as $\mathbf{x}_t=\sqrt{\bar{\alpha}_t} \mathbf{x}_0+\sqrt{1-\bar{\alpha}_t} \boldsymbol{\epsilon}_t$, where $\alpha_t = 1 - \beta_t$, $\bar{\alpha}_t=\prod_{i=0}^t \alpha_i$, and $\boldsymbol{\epsilon}_t \sim \mathcal{N}(\mathbf{0},\mathbf{I})$. 

\subtitle{Reverse Diffusion.} The reverse diffusion process learns to gradually denoise a randomly sampled Gaussian noise $\mathbf{\hat{x}}_{T}\sim\mathcal{N}(\mathbf{0}, \mathbf{I})$ into a sharp result $\mathbf{\hat{x}}_{0}$ conforming to the target sRGB distribution. To strengthen the controllability of the generation procedure, we apply the conditional mechanism~\cite{conditional_ddpm} to improve the fidelity of the reconstructed results conditioned on the amplified RAW feature $\hat{\mathcal{F}}_{raw}$, denoted as $\Tilde{\mathbf{x}}$. The reverse diffusion process can be formulated as:
\begin{equation}\label{eq: reverse_diffusion}
    p_{\theta}(\hat{\mathbf{x}}_{t-1}|\hat{\mathbf{x}}_{t},\widetilde{\mathbf{x}})=\mathcal{N}(\hat{\mathbf{x}}_{t-1};\boldsymbol{\mu}_{\theta}(\hat{\mathbf{x}}_{t},\widetilde{\mathbf{x}},t),\sigma_{t}^{2}\mathbf{I}),
\end{equation}
where $\sigma_{t}^{2}=\frac{1-\overline{\alpha}_{t-1}}{1-\overline{\alpha}_{t}}\beta_{t}$ is the variance and $\boldsymbol{\mu}_{\theta}(\hat{\mathbf{x}}_{t},\widetilde{\mathbf{x}},t)=\frac{1}{\sqrt{\alpha_{t}}}(\hat{\mathbf{x}}_{t}-\frac{\beta_{t}}{\sqrt{1-\overline{\alpha}_{t}}}\boldsymbol{\epsilon}_{\theta}(\hat{\mathbf{x}}_{t},\widetilde{\mathbf{x}},t))$ is the mean value. 

In the training phase, instead of optimizing the parameters of the network $\boldsymbol{\epsilon}_{\theta}$ to promote the estimated noise vector close to Gaussian noise, we follow~\cite{Human_moton_diffusion, FlowDiffuser, DMHomo} to generate the clean sRGB feature $\mathbf{\hat{x}}_{0} = \boldsymbol{\epsilon}_{\theta}(\mathbf{x}_{t}, t, \Tilde{\mathbf{x}})$, i.e., $\hat{\mathcal{F}}_{rgb}$, and employ the content diffusion loss $\mathcal{L}_{cdl}$ for optimization as:
\begin{equation}\label{eq: cdl}
   \mathcal{L}_{cdl}=\mathbb{E}_{\mathbf{x}_0 \sim q\left(\mathbf{x}_0 \mid \tilde{\mathbf{x}}\right), t \sim[1, T]}\left[\left\|\hat{\mathbf{x}}_0-\mathbf{x}_0\right\|_1\right].
\end{equation}
\begin{table*}[!t]
  \centering
  \caption{Quantitative comparisons on the Canon and Sony subsets of the proposed SIED dataset, where the low-light RAW images are with three relatively precise illuminance ranges. The best results are highlighted in \textbf{bold} and the second-best results are in \underline{underlined}. We retrain all comparison methods on the training set of each subset using their officially released codes for fair comparison.}
  \resizebox{\linewidth}{!}{
    \begin{tabular}{c|l|c|ccc|ccc|ccc}
    \toprule
    \multicolumn{12}{c}{\textbf{Canon subset}} \\
    \midrule
    \multirow{2}[3]{*}{Category} & \multirow{2}[3]{*}{Method} & \multirow{2}[3]{*}{Reference} & \multicolumn{3}{c|}{0.01 lux-0.1 lux} & \multicolumn{3}{c|}{0.001 lux-0.01 lux} & \multicolumn{3}{c}{0.0001 lux-0.001 lux} \\
    \cmidrule{4-12} & & & PSNR $\uparrow$ & SSIM $\uparrow$ & LPIPS $\downarrow$ & PSNR $\uparrow$ & SSIM $\uparrow$ & LPIPS $\downarrow$ & PSNR $\uparrow$ & SSIM $\uparrow$ & LPIPS $\downarrow$ \\
    \midrule
    \multirow{5}[0]{*}{Single-stage} 
    & SID~\cite{SID}   & CVPR' 18 &  20.69 & 0.811 & 0.428 & 20.34 & 0.799 & 0.450 & 19.28 & 0.764 & \underline{0.497} \\
    & DID~\cite{DID} & ICME' 19 & 20.34 & 0.806 & 0.422 & 20.05 & 0.798 & \underline{0.447} & 18.84 & 0.760 & 0.520 \\
    & SGN~\cite{SGN}   & ICCV' 19 & 21.79 & 0.813 & \underline{0.421} & 21.07 & \underline{0.800} & \underline{0.447} & 19.42 & 0.762 & 0.514 \\
    & LLPackNet~\cite{LLPackNet} & BMVC' 20 & 20.64 & 0.770 & 0.562 & 20.16 & 0.764 & 0.574 & 18.91 & 0.739 & 0.613 \\
    & RRT~\cite{RRT}   & CVPR' 21 & 20.13 & 0.766 & 0.506 & 19.76 & 0.753 & 0.539 & 18.40 & 0.727 & 0.591 \\
    \midrule
    \multirow{5}[1]{*}{Multi-stage} 
    & LDC~\cite{LDC}   & CVPR' 20 & 20.51 & 0.785 & 0.537 & 19.97 & 0.741 & 0.625 & 18.93 & 0.720 & 0.666 \\
    & MCR~\cite{MCR}   & CVPR' 22 & 21.76 & 0.811 & 0.445 & 21.60 & 0.796 & 0.474 & 19.87 & 0.763 & 0.531 \\
    & DNF~\cite{DNF}   & CVPR' 23 & \underline{24.03} & \underline{0.813} & 0.456 & \underline{23.47} & 0.796 & 0.486 & \underline{21.63} & \underline{0.769} & 0.522 \\
    & RAWMamba~\cite{RAWMamba} & arXiv' 24 & 22.63 & 0.791 & 0.461 & 21.99 & 0.782 & 0.482 & 21.05 & 0.757 & 0.521 \\
    & Ours  & - & \textbf{24.85} & \textbf{0.849} & \textbf{0.360} & \textbf{24.02} & \textbf{0.839} & \textbf{0.379} & \textbf{22.52} & \textbf{0.811} & \textbf{0.435} \\
    \midrule
    \midrule
    \multicolumn{12}{c}{\textbf{Sony subset}} \\
    \midrule
    \multirow{2}[3]{*}{Category} & \multirow{2}[3]{*}{Method} & \multirow{2}[3]{*}{Reference} & \multicolumn{3}{c|}{0.01 lux-0.1 lux} & \multicolumn{3}{c|}{0.001 lux-0.01 lux} & \multicolumn{3}{c}{0.0001 lux-0.001 lux} \\
    \cmidrule{4-12} & & & PSNR $\uparrow$ & SSIM $\uparrow$ & LPIPS $\downarrow$ & PSNR $\uparrow$ & SSIM $\uparrow$ & LPIPS $\downarrow$ & PSNR $\uparrow$ & SSIM $\uparrow$ & LPIPS $\downarrow$ \\
    \midrule
    \multirow{5}[0]{*}{Single-stage} 
    & SID~\cite{SID}   & CVPR' 18 & 21.93 & 0.811 & 0.471 & 21.68 & 0.804 & \underline{0.483} & 20.23 & 0.774 & \underline{0.514} \\
    & DID~\cite{DID} & ICME' 19 &  21.64 & 0.802 & 0.478 & 21.24 & 0.788 & 0.502 & 20.46 & 0.767 & 0.532 \\
    & SGN~\cite{SGN}   & ICCV' 19 & 22.41 & \underline{0.815} & \underline{0.466} & 22.13 & 0.806 & \underline{0.483} & 20.77 & 0.781 & 0.523 \\
    & LLPackNet~\cite{LLPackNet} & BMVC' 20 & 21.57 & 0.787 & 0.554 & 21.37 & 0.782 & 0.568 & 20.33 & 0.763 & 0.596 \\
    & RRT~\cite{RRT}   & CVPR' 21 &  21.11 & 0.773 & 0.540 & 20.95 & 0.777 & 0.552 & 20.49 & 0.756 & 0.600 \\
    \midrule
    \multirow{5}[1]{*}{Multi-stage} 
    & LDC~\cite{LDC}   & CVPR' 20 & 21.74 & 0.797 & 0.517 & 21.20 & 0.785 & 0.555 & 19.59 & 0.749 & 0.637 \\
    & MCR~\cite{MCR}   & CVPR' 22 & 22.87 & 0.810 & 0.492 & 22.26 & 0.806 & 0.502 & 20.95 & 0.771 & 0.542 \\
    & DNF~\cite{DNF}   & CVPR' 23 & \underline{24.24} & 0.814 & 0.473 & \underline{23.91} & \underline{0.807} & 0.497 & \underline{22.42} & \underline{0.785} & 0.531 \\
    & RAWMamba~\cite{RAWMamba} & arXiv' 24 & 23.94 & 0.805 & 0.489 & 23.64 & 0.801 & 0.499 & 21.88 & 0.771 & 0.548 \\
    & Ours  & - &  \textbf{24.98} & \textbf{0.837} & \textbf{0.425} & \textbf{24.51} & \textbf{0.830} & \textbf{0.433} & \textbf{23.10} & \textbf{0.814} & \textbf{0.466} \\
    \bottomrule
    \end{tabular}}
  \label{tab: compare_ours}%
\end{table*}%

\subtitle{Color Consistency Loss.} The color information in RAW images is presented by unique color arrangements within a single channel, which makes the RAW to sRGB conversion potentially vulnerable to unstable color mapping. To this end, we further propose a color consistency loss $\mathcal{L}_{ccl}$ that optimizes the color histogram~\cite{color_histogram} of the generated sRGB feature denoted as $\mathcal{H}_{\hat{\mathcal{F}}_{rgb}}$ to align with the counterpart of the encoded reference sRGB feature as $\mathcal{H}_{\mathcal{F}_{rgb}}$, aiming to facilitate the diffusion model to generate reconstructed sRGB features with vivid color. We employ the KL divergence to quantify the distribution disparity instead of directly minimizing the difference between the $\mathcal{H}_{\hat{\mathcal{F}}_{rgb}}$ and $\mathcal{H}_{\mathcal{F}_{rgb}}$, since the color histogram primarily captures the proportion of various colors across the entire image without accounting for spatial arrangement. Thus, the $\mathcal{L}_{ccl}$ is formulated as:
\begin{equation}\label{eq: ccl}
   \mathcal{L}_{ccl}=\sum_{c \in [0, C)} \mathcal{H}_{\hat{\mathcal{F}}_{rgb}^{c}} \log (\frac{\mathcal{H}_{\hat{\mathcal{F}}_{rgb}^{c}}}{\mathcal{H}_{\mathcal{F}_{rgb}^{c}} + \tau}),
\end{equation}
where $\tau$ is a small constant to avoid the zero denominator.

In the inference phase, we derive the restored sRGB feature $\hat{\mathcal{F}}_{rgb}$ from learned distribution through the reverse diffusion process following~\cite{ddim}, and then send it to the sRGB decoder $\mathcal{D}_{rgb}(\cdot)$ to produce the final sRGB image $\hat{I}_{rgb}$.

\subsection{Network Training}\label{subsec: network_training}
Our approach employs a two-stage training strategy. In the first stage, we use the paired low-light RAW images and reference normal-light RAW/sRGB images to optimize the RAW encoder-decoder ($\mathcal{E}_{raw}$ and $\mathcal{D}_{raw}$), sRGB encoder-decoder ($\mathcal{E}_{rgb}$ and $\mathcal{D}_{rgb}$), and AICM, while freezing the parameters of the diffusion model. The encoders and decoders are optimized with the content loss $\mathcal{L}_{con}$ as:
\begin{equation}
    \mathcal{L}_{con} = \sum_{i = \{raw, rgb\}} ||I_{i} - \mathcal{D}_{i}(\mathcal{E}_{i}(I_{i}))||_1.
\end{equation}
The optimization objective in the first stage is formulated as $\mathcal{L}_{stage1} = \mathcal{L}_{con} + \mathcal{L}_{icl}$. In the second stage, we optimize the diffusion model through $\mathcal{L}_{stage2} = \mathcal{L}_{cdl} + \lambda \mathcal{L}_{ccl}$, while freezing the parameters of other modules. 
\begin{figure*}[!t]
    \centering
    \includegraphics[width=\linewidth]{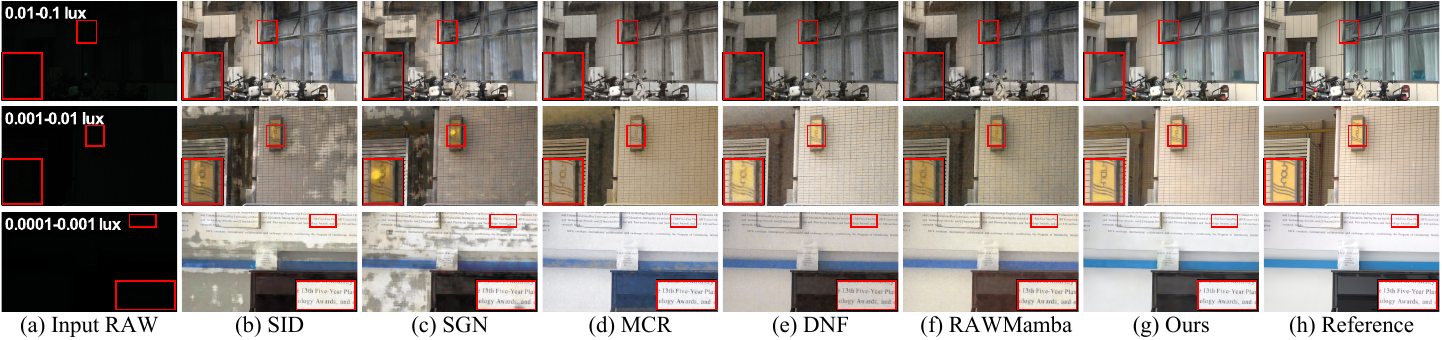}
    \caption{Qualitative comparison of our method and competitive methods~\cite{SID,SGN,MCR,DNF,RAWMamba} on the Canon subset of the proposed SIED dataset. 
    The input RAW images are visualized by a simple demosaicing process. Best viewed by zooming in.}
    \label{fig: Visual_canon}
\end{figure*}
\begin{figure*}[!t]
    \centering
    \includegraphics[width=\linewidth]{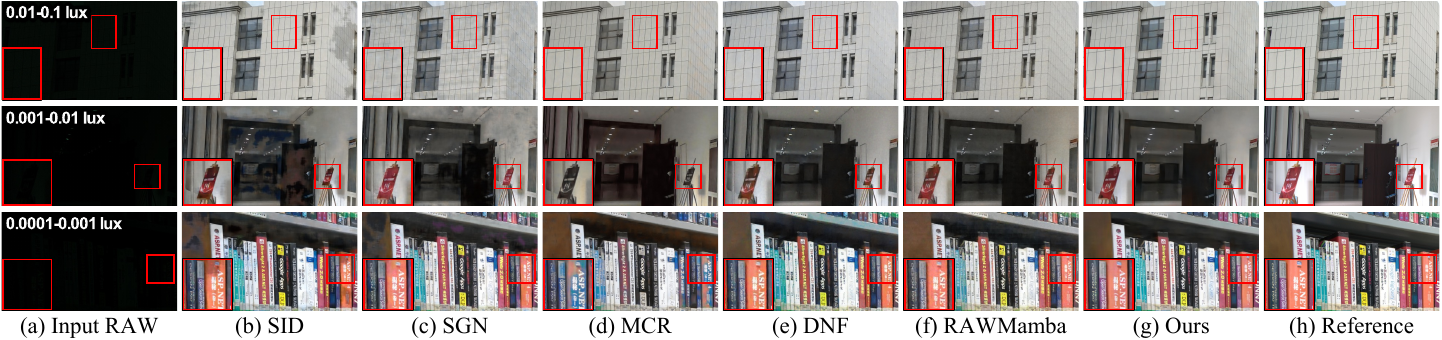}
    \caption{Qualitative comparison of our method and competitive methods~\cite{SID,SGN,MCR,DNF,RAWMamba} on the Sony subset of the proposed SIED dataset. The input RAW images are visualized by a simple demosaicing process. Best viewed by zooming in.}
    \label{fig: Visual_sony}
\end{figure*}

\section{Experiments}\label{sec: experiments}
\subsection{Experimental Settings}\label{subsec: experimental_settings}
\subtitle{Implementation Details.} We implement the proposed method with PyTorch on one NVIDIA A100 GPU, where the batch size and patch size are set to 1 and 512 $\times$ 512. The networks can be converged after training in two stages with $2 \times 10^5$ and $4 \times 10^5$ iterations, respectively. We employ the Adam optimizer~\cite{Adam} for optimization with the initial learning rate set to $1 \times 10^{-4}$ in the first stage and decays by a factor of 0.8 while reinitializing it to a fixed value of $8 \times 10^{-5}$ in the second stage. The hyper-parameter $\lambda$ is set to 0.1. For the diffusion model, we adopt the U-Net~\cite{Unet} architecture as the noise estimator network with the time step and sampling step set to 1000 and 20, respectively.

\subtitle{Datasets and Metrics.} To evaluate the performance of our method, we conducted experiments on the Canon and Sony subsets of our proposed SIED dataset, which contain low-light RAW images under three different illumination conditions, i.e., 0.01-0.1 lux, 0.001-0.01 lux, and 0.0001-0.001 lux. Moreover, we also conduct experiments on the SID~\cite{SID} dataset which contains low-light RAW images with normal-light reference images captured by Sony and Fuji cameras. Two distortion metrics PSNR and SSIM~\cite{SSIM}, and a perceptual metric LPIPS~\cite{LPIPS} are adopted for evaluation.
\begin{table}[!t]
  \centering
  \caption{Quantitative comparisons on the Sony and Fuji subsets of SID~\cite{SID} dataset. The best results are highlighted in \textbf{bold} and the second-best results are in \underline{underlined}. ``-" indicates the result is not available since the pre-trained model has not been released.}
  \LARGE
  \resizebox{\linewidth}{!}{
    \begin{tabular}{c|l|ccc|ccc}
    \toprule
    \multirow{2}[4]{*}{} & \multirow{2}[4]{*}{Method} & \multicolumn{3}{c|}{Sony} & \multicolumn{3}{c}{Fuji} \\
    \cmidrule{3-8}          &       & PSNR $\uparrow$ & SSIM $\uparrow$ & LPIPS $\downarrow$ & PSNR $\uparrow$ & SSIM $\uparrow$ & LPIPS $\downarrow$ \\
    \midrule
    \multirow{5}[2]{*}{\rotatebox{90}{Sing-stage}} 
    & SID~\cite{SID}  & 28.96 & 0.787 & 0.356 & 26.66 & 0.709 & 0.432 \\
    & DID~\cite{DID}   & 29.16 & 0.785 & 0.368 & - & - & - \\
    & SGN~\cite{SGN}   & 29.28 & 0.790 & 0.370 & 27.41 & 0.720 & 0.430 \\
    & LLPackNet~\cite{LLPackNet} & 27.83 & 0.755 & 0.541 & - & - & - \\
    & RRT~\cite{RRT}   & 28.66 & 0.790 & 0.397 & 26.94 & 0.712 & 0.446 \\
    \midrule
    \multirow{5}[2]{*}{\rotatebox{90}{Mlti-stage}} 
    & LDC~\cite{LDC}   & 29.56 & \underline{0.799} & 0.359 & 27.18 & 0.703 & 0.446 \\
    & MCR~\cite{MCR}   & 29.65 & 0.797 & 0.348 & - & - & - \\
    & DNF~\cite{DNF} & \underline{30.62} & 0.797 & \underline{0.343} & \underline{28.71} & \textbf{0.726} & \underline{0.391} \\
    & RAWMamba~\cite{RAWMamba} & \underline{30.62} & 0.794 & 0.350 & 28.65 & \underline{0.723} & 0.460 \\
    & Ours  & \textbf{31.20} & \textbf{0.801} & \textbf{0.339} & \textbf{28.90} & \textbf{0.726} & \textbf{0.390} \\
    \bottomrule
    \end{tabular}}
  \label{tab: comapre_sid}%
\end{table}%

\subsection{Comparison with Existing Methods}\label{subsec: comparisons}
\subtitle{Comparison Methods.} We compare our method with existing RAW-based LLIE methods including single-stage methods SID~\cite{SID}, DID~\cite{DID}, SGN~\cite{DID}, LLPackNet~\cite{LLPackNet}, and RRT~\cite{RRT}, as well as multi-stage methods LDC~\cite{LDC}, MCR~\cite{MCR}, DNF~\cite{DNF}, and RAWMamba~\cite{RAWMamba}. For our SIED dataset, we retrain all comparison methods on the training set of each subset for fair comparison. For the SID dataset, we retrain our method on it and adopt the released pre-trained weights of the comparison methods for evaluation. 

\subtitle{Quantitative Comparison.} We first compare the proposed method with all comparison methods on the Canon and Sony subsets of our proposed SIED dataset. As shown in Table~\ref{tab: compare_ours}, multi-stage methods typically present superior performance than single-stage methods, where our method achieves state-of-the-art performance in terms of both distortion metrics and perceptual metrics across various illumination conditions. To further validate the effectiveness of our method, we also compare our method with comparison methods on the Sony and Fuji subsets of the SID~\cite{SID} dataset. As shown in Table~\ref{tab: comapre_sid}, our method also outperforms previous competitors in all metrics. It is worth noting that, on the SID dataset, previous methods typically use the GT exposure to pre-amplify the low-light RAW image, while our method estimates the amplification coefficients within the proposed AICM, further proving the effectiveness of our method.
\begin{figure}[!t]
    \centering
    \includegraphics[width=\linewidth]{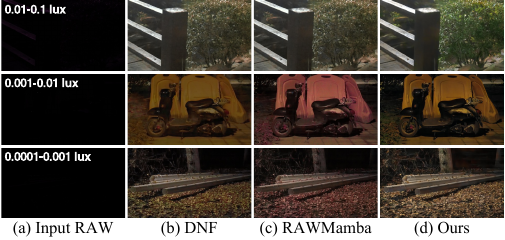}
    \caption{Qualitative comparison of our method and competitive methods~\cite{DNF, RAWMamba} on real-world extremely dark scenes.}
    \label{fig: real_data}
\end{figure}

\subtitle{Qualitative Comparison.} We present visual comparisons of our method and competitive methods on the Canon and Sony subsets of our proposed SIED dataset in Fig.~\ref{fig: Visual_canon} and Fig.~\ref{fig: Visual_sony}, where the input RAW images in rows 1-3 are with the illuminance of 0.01-0.1 lux, 0.001-0.01 lux, and 0.0001-0.001 lux, respectively. Single-stage methods suffer from color distortion and unexpected artifacts, while multi-stage competitors present blurred details, color deviation, or noise amplification. In contrast, our method properly improves contrast, reconstructs sharper details, presents vivid color, and suppresses noise, resulting in visually pleasing results. 

\subtitle{Real-world Generalization.} To validate the generalization ability of our method, we collect several images in realistic extremely dark scenes using the Sony camera and determine their illuminance level according to the matching of the illumination histograms in the Y-channel to standard laboratory images. As shown in Fig.~\ref{fig: real_data}, we use our method and two competitive methods~\cite{DNF, RAWMamba} trained on the Sony subset of the proposed SIED dataset to restore the collected realistic images, where all methods can transform extremely low-light images into normal-light counterparts while our method performs better, proving the effectiveness of the proposed method and our synthesized dataset is capable of supporting methods to generalize to real-world scenes.

\subsection{Ablation Studies}\label{subsec: ablations}
In this section, we conduct a series of ablation studies to validate the impact of different component choices. The quantitative results for the 0.01-0.1 lux illuminance in the Canon subset of our SIED dataset are illustrated in Table~\ref{tab: ablation}. 

\subtitle{Training Strategy.} To validate the effectiveness of our adopted multi-stage training strategy, we conduct an experiment by optimizing the encoder-decoders, AICM, and diffusion model simultaneously. As shown in row 1 of Table~\ref{tab: ablation}, the single-stage training strategy causes overall performance degradation since the encoded features in the early training phase are not favorable for the diffusion model to learn desired target distributions. Nevertheless, our method trained with the single-stage strategy also outperforms previous single-stage competitors in Table~\ref{tab: compare_ours}.

\subtitle{Module Effectiveness.} To validate the effectiveness of our proposed AICM, we conduct experiments by removing it from the overall framework as well as replacing it with fixed amplification factors ($\operatorname{Amp.}$) of 100, 200, and 300, respectively. As shown in rows 2-5 of Table~\ref{tab: ablation}, with the adaptive exposure improvement realized by our AICM, our method achieves significant performance gains in terms of PSNR compared to fixed amplification factors since the illumination degradation in realistic scenes is diverse and unknown.

\begin{figure}[!t]
    \centering
    \includegraphics[width=\linewidth]{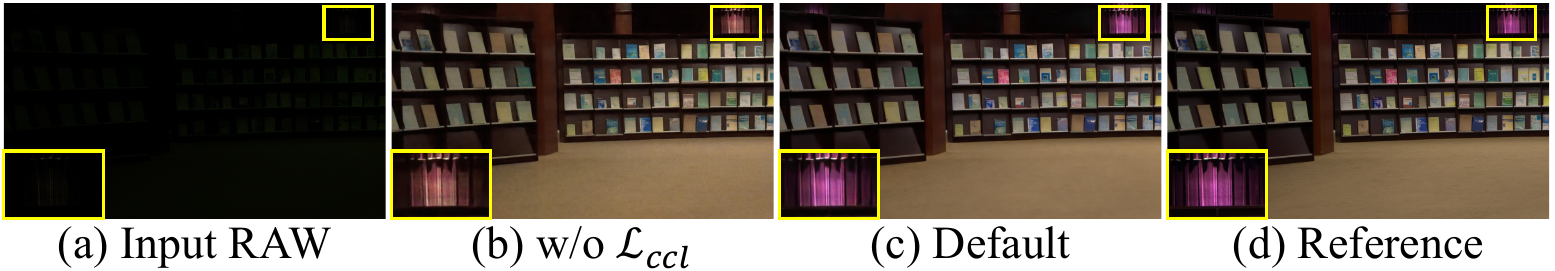}
    \caption{Visual results of the ablation study about our proposed color consistency loss $\mathcal{L}_{ccl}$. `w/o' denotes without.}
    \label{fig: ablation}
\end{figure}

\subtitle{Loss Function.} To validate the effectiveness of the proposed color consistency loss $\mathcal{L}_{ccl}$, we conduct an experiment to remove it from the object function utilized to optimize the diffusion model. As reported in rows 6-7 of Table~\ref{tab: ablation}, the incorporation of $\mathcal{L}_{ccl}$ results in performance superiority especially in terms of structural similarity and perceptual quality. As shown in Fig.~\ref{fig: ablation}, the proposed $\mathcal{L}_{ccl}$ is helpful to achieve accurate color mapping.
\begin{table}[!t]
  \centering
  \caption{Quantitative results of ablation studies, please refer to the text for more details. `w/o' denotes without.}
  \resizebox{\linewidth}{!}{
    \begin{tabular}{r|l|ccc}
    \toprule
    & Method & PSNR $\uparrow$ & SSIM $\uparrow$ & LPIPS $\downarrow$ \\
    \midrule
    1) & Single-stage & 22.96 $_{\textcolor{gray}{(-1.89)}}$ & 0.809 $_{\textcolor{gray}{(-0.040)}}$ & 0.431 $_{\textcolor{gray}{(+0.071)}}$ \\
    \midrule
    2) & w/o AICM & 23.23 $_{\textcolor{gray}{(-1.62)}}$ & 0.839 $_{\textcolor{gray}{(-0.010)}}$ & 0.378 $_{\textcolor{gray}{(+0.018)}}$ \\
    3) & $\operatorname{Amp.}  = 100$ &  23.74 $_{\textcolor{gray}{(-1.11)}}$ & 0.841 $_{\textcolor{gray}{(-0.008)}}$ & 0.373 $_{\textcolor{gray}{(+0.013)}}$ \\
    4) & $\operatorname{Amp.}  = 200$ &  23.48 $_{\textcolor{gray}{(-1.37)}}$ & 0.844 $_{\textcolor{gray}{(-0.005)}}$ & 0.371 $_{\textcolor{gray}{(+0.011)}}$ \\
    5) & $\operatorname{Amp.}  = 300$ & 23.18 $_{\textcolor{gray}{(-1.67)}}$ & 0.838 $_{\textcolor{gray}{(-0.011)}}$ & 0.382 $_{\textcolor{gray}{(+0.022)}}$ \\
    \midrule
    6) & w/o $\mathcal{L}_{ccl}$ & 24.53 $_{\textcolor{gray}{(-0.32)}}$ & 0.836 $_{\textcolor{gray}{(-0.013)}}$ & 0.389 $_{\textcolor{gray}{(+0.029)}}$ \\
    \midrule
    7) & default & 24.85 $_{\textcolor{gray}{(+0.00)}}$ & 0.849 $_{\textcolor{gray}{(+0.000)}}$ & 0.360 $_{\textcolor{gray}{(+0.000)}}$ \\
    \bottomrule
    \end{tabular}}
  \label{tab: ablation}
\end{table}

\vspace{1mm}
\section{Conclusion}\label{sec: conclusion}
We have presented a paired-to-paired data synthesis pipeline to prepare a more challenging dataset named SIED for extremely low-light RAW image enhancement, containing low-light RAW images at three well-calibrated illuminance levels as low as 0.0001 lux, along with high-quality reference sRGB images. Moreover, we propose a diffusion-based multi-stage framework that leverages the generative ability and denoising property of diffusion models to restore visually pleasing results from extremely low-light inputs. Technically, we propose an adaptive illumination correction module that performs illumination pre-amplification within the latent space, aiming to obtain better results and avoid exposure bias in diffusion processes. Moreover, we propose a color consistency loss based on the color histogram to promote the diffusion model to generate reconstructed sRGB images with accurate color mapping. Experimental results demonstrate the superiority of our method. 

\noindent\textbf{Acknowledgements.} This work was supported by the National Natural Science Foundation of China (NSFC) under grants No.62372091 and No.62301310.

\clearpage
{\small
    \bibliographystyle{ieeenat_fullname}
    \bibliography{main}}

\end{document}